\documentclass[letterpaper, 10 pt, conference]{ieeeconf}

\IEEEoverridecommandlockouts
\usepackage{graphicx}
\usepackage{amsmath}
\usepackage{amssymb}
\usepackage{booktabs}
\usepackage{multirow}
\usepackage{cite}
\usepackage{balance}
\usepackage{xurl}
\usepackage[hidelinks]{hyperref}
\usepackage{algorithm}
\usepackage{algpseudocode}
\usepackage[font=small]{caption}

\title{\LARGE \bf
Action Agent: Agentic Video Generation Meets Flow-Constrained Diffusion
}

\author{
Jeffrin Sam*,
Nguyen Khang*,
Yara Mahmoud,
Miguel Altamirano Cabrera,
Dzmitry Tsetserukou%
\thanks{* Denotes equal contribution.}
\thanks{Authors are with the Intelligent Space Robotics Laboratory, Skoltech,
         Bolshoy Boulevard, 30, Moscow 121205, Russia. %
{\tt\small \{Jeffrin.Sam, Nguyen.Khang, Yara Mahmoud, M.Altamirano, D.Tsetserukou\}@skoltech.ru}}}

\begin{document}

\maketitle
\thispagestyle{empty}
\pagestyle{empty}


\begin{abstract}

We present \textbf{Action Agent}, a two-stage framework that unifies agentic navigation video generation with flow-constrained diffusion control for multi-embodiment robot navigation. In Stage~I, a large language model~(LLM) acts as an orchestration module that selects video diffusion models, refines prompts through iterative validation, and accumulates cross-task memory to synthesize physically plausible first-person navigation videos from language and image inputs. This increases video generation success from 35\% (single-shot) to 86\% across 50 navigation tasks. In Stage~II, we introduce \textbf{FlowDiT}, a Flow-Constrained Diffusion Transformer that converts optimized goal videos and language instructions into continuous velocity commands using action-space denoising diffusion. FlowDiT integrates DINOv2 visual features, learned optical flow for ego-motion representation, and CLIP language embeddings for semantic stopping. We pretrain on the RECON outdoor navigation dataset and fine-tune on 203 Unitree G1 humanoid episodes collected in Isaac Sim to calibrate velocity dynamics. A single 43M-parameter checkpoint achieves 73.2\% navigation success in simulation and 64.7\% task completion on a real Unitree G1 in unseen indoor environments under open-loop execution, while operating at 40--47\,Hz. We evaluate Action Agent across three embodiments: a Unitree G1 humanoid (real hardware), a drone, and a wheeled mobile robot (Isaac Sim), demonstrating that decoupling trajectory imagination from execution yields a scalable and embodiment-aware paradigm for language-guided navigation.

\end{abstract}

\section{Introduction}

\begin{figure}[t]
    \centering
    \includegraphics[width=0.9\columnwidth]{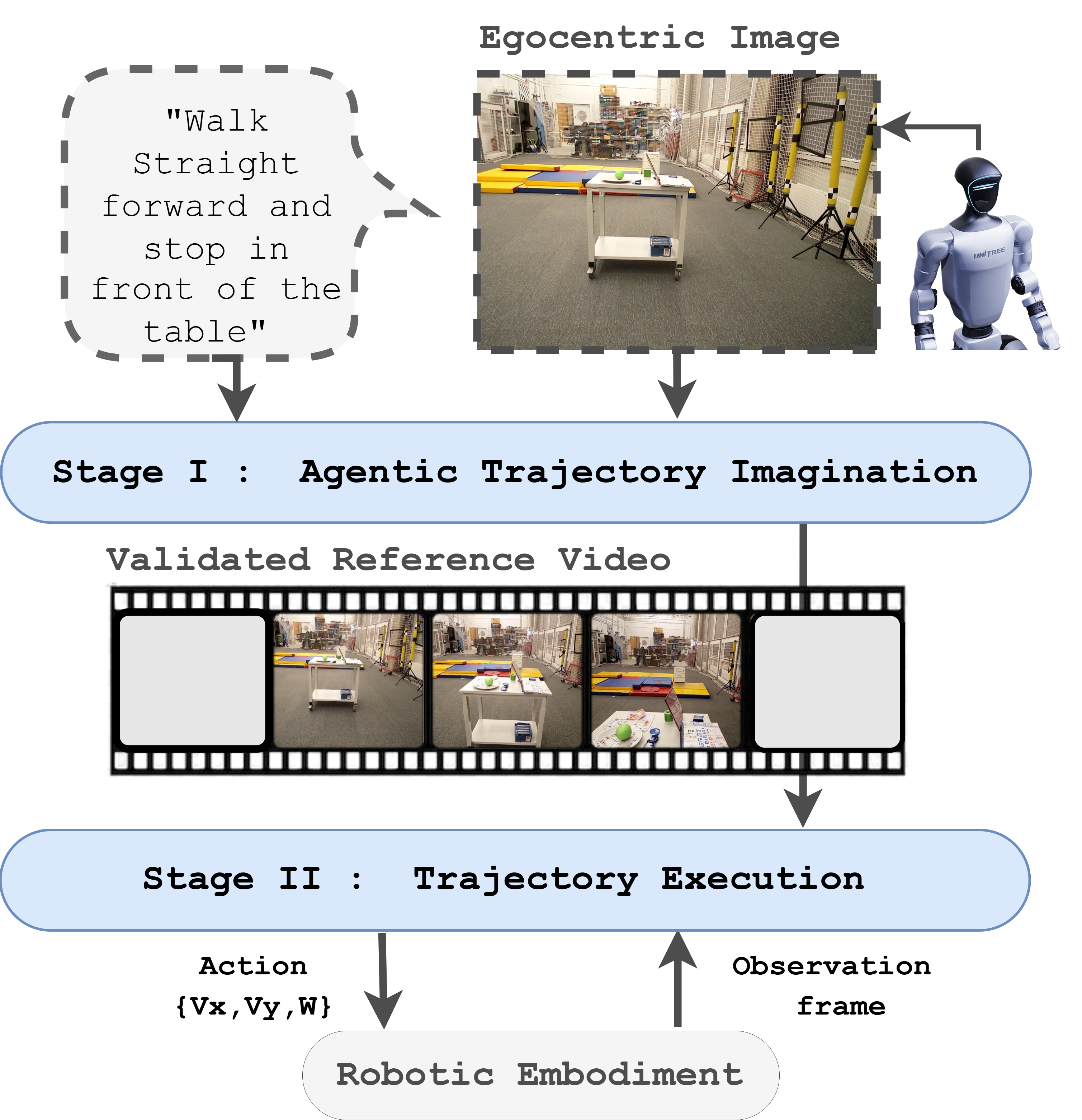}
    \caption{Action Agent system overview. Stage I performs agentic trajectory imagination by generating and validating a first-person \emph{visual intermediate representation} (reference navigation video) from a language instruction and an initial observation. Stage II executes this reference using FlowDiT to produce continuous velocity commands for deployment across multiple robot embodiments.
    }
    \label{fig:sys_teaser}
    \vspace{-0.8cm}
\end{figure}

Robot navigation from high-level natural language instructions remains a fundamental challenge in embodied AI. Bridging language to reliable motion requires interpreting task intent, grounding it in a specific scene, and producing physically plausible behavior under diverse environments and robot embodiments.

Classical navigation pipelines address this complexity through a modular decomposition: simultaneous localization and mapping~(SLAM) to estimate pose and build a geometric representation, waypoint or trajectory planning over that representation, and hand-engineered controllers to track the plan. This separation offers interpretability and safety hooks, but it often requires careful tuning per robot and sensor suite, and it can degrade under distribution shift (e.g., changes in layout, lighting, clutter, or dynamics) or when transferring to new embodiments~\cite{durrantwhyte_bailey_2006_slam,paden_2016_planning_control_survey}.

Conversely, recent end-to-end visuomotor policies aim to bypass explicit planning by directly mapping observations (and sometimes language) to actions. While effective in narrow regimes, such monolithic Vision-Language-Action~(VLA) approaches can entangle high-level intent with low-level control. This ``black box'' nature makes adaptation and failure recovery difficult, demands immense computational footprints that preclude high-frequency edge deployment, and frequently suffers from spatial reasoning hallucinations~\cite{hussein_2017_imitation_learning_survey,openvla_2024,rtx_2023}.

We propose a two-stage decomposition that makes the interface between \emph{intention} and \emph{execution} explicit. While end-to-end models entangle intent with control, Action Agent introduces an explicit \textbf{Visual Intermediate Representation~(VIR)}. This allows the system to ``rehearse'' the trajectory in the pixel space-where generative foundation models are most capable-before committing to metric-space actions. An overview of the proposed system is shown in Fig.~\ref{fig:sys_teaser}.

\begin{itemize}
    \item \textbf{Stage~I: Digital rehearsal via agentic video generation.}
    Given an instruction and a scene observation, an AI agent orchestrates a vision-language model and a text-to-video generator to synthesize candidate first-person navigation videos. A reasoning module scores each candidate using a structured rubric (e.g., instruction adherence, physics consistency). The agent iteratively refines prompts using this feedback and a cross-task memory buffer until a physically plausible VIR is obtained~\cite{stable_video_diffusion_2023,cosmos_2025}.

    \item \textbf{Stage~II: Diffusion-based trajectory execution via FlowDiT.}
    Conditioned on the validated VIR and language -and optionally live observations- FlowDiT produces continuous velocity commands $(v_x, v_y, \omega)$. It supports both open-loop execution (from the reference video alone) and closed-loop operation at 40--47\,Hz with 43M trainable parameters.
\end{itemize}

This decomposition enables several practical capabilities:
\begin{itemize}
    \item \textbf{Embodiment-aware trajectory imagination:} Stage~I conditions generation on embodiment-relevant constraints via the agentic validation loop, producing a VIR that respects the robot's physical capabilities.
    \item \textbf{Multi-modal action modeling:} Stage~II integrates appearance-driven cues (RGB semantics) with explicit local geometry (optical flow) to resolve ``looming'' depth ambiguities~\cite{dinov2_2023, raft_2020}.
    \item \textbf{Language-conditioned stopping:} The VIR encodes motion, while language explicitly guides semantic termination (where/when to stop)~\cite{clip_2021,saycan_2022}.
    \item \textbf{Edge-deployable execution:} FlowDiT achieves a $161\times$ parameter reduction compared to state-of-the-art VLAs, supporting high-frequency (40-47\,Hz) reactive execution on consumer hardware.
\end{itemize}

In this paper, we evaluate \textbf{Stage~I} and report how agentic digital rehearsal improves the reliability of synthesized navigation videos. We then evaluate \textbf{Stage~II} across three embodiments: a wheeled mobile robot, a drone, and open-loop execution on real Unitree G1 humanoid hardware. The goal is to demonstrate how explicitly separating visual imagination from metric control creates a scalable pathway for foundation models in robotics.

\section{Related Work}

\subsection{Diffusion Models for Control and Navigation}
Diffusion models have emerged as a powerful tool for robotics control due to their ability to represent multi-modal action distributions and to generate temporally coherent trajectories~\cite{ddpm_2015,ddim_2021,dit_2023}.
In robotics, diffusion-based policies have been applied to manipulation and control as an alternative to regression objectives that can average across distinct valid behaviors~\cite{diffusion_policy_2023}.
In navigation contexts, diffusion formulations have also been used for goal conditioning and sequential decision-making, including video- or goal-conditioned variants such as NoMaD~\cite{nomad_2024}. This literature motivates diffusion-based execution modules that can generate action sequences conditioned on rich context while maintaining robustness to multi-modality and ambiguity. However, such approaches typically rely on geometric or latent goal representations rather than explicitly imagined visual trajectory references.

\subsection{Ego-Motion Representations and Optical Flow for Control}
Beyond semantic visual representations, ego-motion cues can provide complementary information for control, especially for short-horizon motion and directional disambiguation.
Optical flow has long been studied as a motion representation in visual odometry and SLAM pipelines, and as a cue for camera motion estimation~\cite{raft_2020,horn_schunck_1981,lucas_kanade_1981}.
In learning-based navigation and control, explicit motion representations (e.g., flow or ego-motion predictors) can reduce sensitivity to appearance variation and provide a more direct signal for local motion dynamics~\cite{flowcontrol_2020,katara_2021_dmpc_flow}.
In parallel, generative video modeling research has explored imposing flow- or motion-related constraints to improve temporal consistency of synthesized videos~\cite{motionctrl_2024,liang_2024_motionaware_video_diffusion}.
These works collectively suggest that incorporating explicit ego-motion representations alongside semantic features is a promising direction for stabilizing control under changing visual conditions.

\subsection{Language-Conditioned Vision-Based Navigation}
Language-conditioned navigation has been approached through both classical modular stacks and learning-based policies.
Classical pipelines typically factor navigation into (i)~mapping and localization (e.g., SLAM), (ii)~explicit planning over a geometric representation, and (iii)~low-level control for plan tracking~\cite{durrantwhyte_bailey_2006_slam,paden_2016_planning_control_survey}.
This decomposition provides interpretability and safety hooks, but often requires careful tuning and can degrade under distribution shift or when transferring across embodiments (e.g., viewpoint, sensor suite, and dynamics)~\cite{thrun_2005_probabilistic_robotics,durrantwhyte_bailey_2006_slam}.
Learning-based navigation methods instead train visuomotor policies that map observations (and sometimes language) to actions, aiming to reduce reliance on explicit mapping and planning~\cite{vint_2023,nomad_2024}. Vision-based transformer policies such as ViNT~\cite{vint_2023} adopt pretrained visual encoders to directly map RGB observations to navigation actions, demonstrating strong generalization across indoor scenes. Nevertheless, these models entangle high-level intent with low-level control, without explicitly separating trajectory imagination from execution.

\subsection{Video-Conditioned Navigation and Learning from Demonstrations}
A growing body of work studies learning robot policies from demonstrations represented as trajectories, videos, or video-like references.
Imitation learning and learning-from-demonstration approaches commonly rely on expert trajectories collected in simulation or on real platforms~\cite{hussein_2017_imitation_learning_survey,il_survey_2024}.
More recent methods explore using \emph{videos} as an intermediate representation for goal conditioning or policy learning.
For example, diffusion-based policy learning has been used to model action distributions from demonstrations~\cite{diffusion_policy_2023}, and goal-conditioned navigation systems have been proposed that condition policies on reference videos to specify intended behavior~\cite{nomad_2024}.
Other approaches unify multiple sensing and action modalities through shared video representations or latent video objectives~\cite{unipi_2024}.
These lines of work motivate the use of video as a compact interface between intent and control, while also highlighting the dependence of many approaches on the availability and quality of reference demonstrations.

\subsection{LLMs for Robotics and Agentic Iterative Refinement}
Recent advances in video diffusion models (e.g., Stable Video Diffusion~\cite{stable_video_diffusion_2023}, Runway Gen-2~\cite{runway_gen2_2023}, Cosmos~\cite{cosmos_2025}) enable high-quality task-conditioned video generation. However, naive prompting often produces videos that violate physical constraints or fail to align with robot embodiment capabilities.
Large language models have been used for robotic reasoning in works such as RT-X~\cite{rtx_2023}, Gato~\cite{gato_2022}, Code-as-Policies~\cite{code_as_policies_2022}, and SayCan~\cite{saycan_2022}. These methods typically apply LLMs in forward planning or few-shot reasoning settings.
In contrast, we treat the LLM as a meta-optimizer that iteratively refines generation prompts using structured multi-objective feedback~\cite{self_refine_2023,inner_monologue_2022,progprompt_2023}.
Such rubric-based validation aligns with recent ``LLM-as-a-judge'' evaluation frameworks that approximate human preference with strong model-based scoring~\cite{llm_judge_mtbench_2023}.
More recently, large-scale vision-language-action models such as OpenVLA~\cite{openvla_2024} leverage foundation visual encoders and large language models to directly produce robotic actions from multimodal inputs. While these models benefit from scale, they often require substantial trainable capacity and extensive pretraining, motivating more parameter-efficient alternatives for embodiment-aware navigation.

\begin{figure*}[t]
    \centering
    \vspace{0.4cm}
    \includegraphics[width=0.90\textwidth]{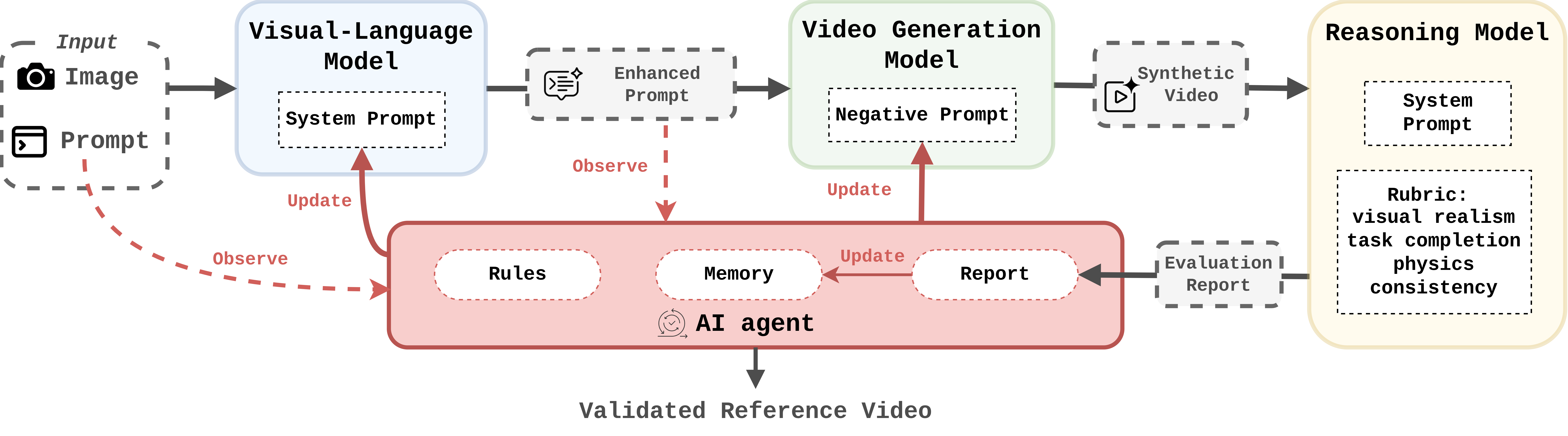}
    \caption{Stage~I: agentic trajectory imagination (digital rehearsal). A central LLM agent orchestrates (i) a vision--language model for prompt construction, (ii) a text-to-video diffusion generator, and (iii) a reasoning-based evaluator that scores candidate videos and produces structured critiques. The agent iteratively refines prompts using feedback and memory until a validated reference navigation video $V_{\mathrm{goal}}$ is obtained.}
    \label{fig:stage1}
    \vspace{-0.5cm}
\end{figure*}

\section{Method}

Action Agent decomposes language-guided navigation into two distinct stages to explicitly decouple high-level \textit{trajectory imagination} from low-level \textit{control execution}:
\begin{align}
\textbf{Stage~I:} \quad & (L, I_0) \rightarrow V_{goal} \\
\textbf{Stage~II:} \quad & (V_{goal}, [I_t], L) \rightarrow a_t,
\end{align}
where $L$ is a natural language instruction, $I_0$ is the initial scene observation, $V_{goal}$ is the synthesized reference video, $I_t$ is an optional live camera feed (brackets denote optionality), and $a_t$ is a sequence of continuous velocity commands. When $I_t$ is unavailable, FlowDiT operates in \emph{open-loop} mode using $V_{goal}$ and $L$ alone; when available, it enables closed-loop reactive execution.

The detailed per-task optimization procedure is summarized in
Algorithm~\ref{alg:agentic_loop}.

\begin{algorithm}[t]
\vspace{0.2cm}
\centering
\fbox{%
\begin{minipage}{0.96\columnwidth}
\footnotesize
\textbf{LLM-based agentic optimization loop (per task)}\\
\textbf{Input:} language $L$, image(s) $I_{0}$ (and optionally $I_g$), model pool $\mathcal{M}=\{\text{WAN},\text{LTX}\}$\\
\textbf{1.} Load \texttt{rules.md}, long-term memory $M_{LT}$, short-term memory $M_{ST}\leftarrow\varnothing$.\\
\textbf{2.} Select model $m \leftarrow \mathrm{route}(I_0, I_g, M_{LT})$.\\
\textbf{3.} For $iter=1$ to $K_{\max}$:\\
\hspace*{1.2em}\textbf{3.1} $p \leftarrow \mathrm{VLM}(L,I_0; m, M_{LT})$ \hfill(enhanced prompt).\\
\hspace*{1.2em}\textbf{3.2} $V \leftarrow \mathrm{GenVideo}(m,p)$.\\
\hspace*{1.2em}\textbf{3.3} $(\mathrm{PA},\mathrm{PP},\mathrm{VQ}),R \leftarrow \mathrm{Evaluator}(V,L)$.\\
\hspace*{1.2em}\textbf{3.4} If $\mathrm{mean}(\mathrm{PA},\mathrm{PP},\mathrm{VQ}) \ge \tau$, update $M_{LT}$ and return $V$.\\
\hspace*{1.2em}\textbf{3.5} Else set $b \leftarrow \mathrm{identify\_bottleneck}(\mathrm{PA},\mathrm{PP},\mathrm{VQ})$.\\
\hspace*{2.4em}Set $p \leftarrow \mathrm{refine\_prompt}(p,b,R,M_{ST},M_{LT})$.\\
\hspace*{2.4em}Append $(p,b,R)$ to $M_{ST}$.\\
\textbf{4.} Return best-found $V$ (or fail).
\end{minipage}%
}
\caption{LLM-based agentic optimization loop (per task).}
\label{alg:agentic_loop}
\end{algorithm}

\subsection{Stage~I: Agentic Trajectory Imagination as Digital Rehearsal}
Unlike classical planning pipelines that rely on pre-built geometric maps, Stage~I performs trajectory planning directly in the visual domain by synthesizing a first-person reference video $V_{goal}$. However, single-shot generative video frequently suffers from physical hallucinations (e.g., passing through solid objects or violating kinematic constraints).

To bridge generative imagination and physical reality, we frame Stage~I as a \textbf{rehearsal-based validation layer}. As illustrated in Fig.~\ref{fig:stage1}, a central large language model acts as an orchestration agent that iteratively proposes, generates, and critiques candidate trajectories. It coordinates three components:
(i)~a vision-language model (Qwen3-VL~\cite{qwen3vl_2025}) for structured prompt construction,
(ii)~a text-to-video diffusion generator (WAN~2.2~\cite{wan2_2025} or LTX-Video~\cite{ltxvideo_2024}),
and (iii)~a reasoning-based validator (Cosmos-Reason1~\cite{cosmos_2025}) that evaluates candidate videos against structured physical and visual criteria.

By evaluating these candidates against structured criteria, the agent effectively ``rehearses'' the trajectory in pixel space. This formulation ensures that the generated \emph{Visual Intermediate Representation~(VIR)} is physically plausible \emph{before} any velocity commands are issued to the robot. Consequently, Action Agent trades offline planning compute (the agentic loop) for online execution safety, decoupling semantic reasoning from the high-frequency control loop of Stage~II.

\subsubsection{Formal Objective}
We formulate Stage~I as a discrete optimization over the space of prompt modifiers and pipeline parameters $p \in \mathcal{P}$. The objective is to maximize a multi-modal reward function evaluated by the reasoning module:
\begin{align}
p^{\star} &= \arg\max_{p\in\mathcal{P}} \Big( \lambda_1 \,\text{PA}(V_p) + \lambda_2 \,\text{PP}(V_p) + \lambda_3 \,\text{VQ}(V_p) \Big) \nonumber \\
&\text{subject to} \quad \mathrm{mean}(\text{PA}, \text{PP}, \text{VQ}) \ge \tau,
\end{align}
where $V_p$ is the generated video under strategy $p$, and PA, PP, and VQ denote Prompt Adherence, Physical Plausibility, and Visual Quality, respectively. The threshold $\tau$ (empirically set to 80) enforces minimum safety and quality constraints before the trajectory is accepted for execution.

The search over $\mathcal{P}$ is realized through discrete parameter edits rather than continuous gradient updates. The orchestrating LLM proposes modifications such as: (i)~altering motion descriptors, (ii)~adjusting camera dynamics, (iii)~modifying negative prompts, and (iv)~switching video generators. The search policy is heuristic and bottleneck-first, executing a greedy local update targeting the lowest-scoring rubric component to minimize iterations.

\subsubsection{Agentic Optimization Loop}
The detailed per-task procedure is summarized in Algorithm~\ref{alg:agentic_loop}. For each task, the agent executes the following loop bounded by $K_{\max}=5$:
\begin{itemize}
\item \textbf{Model Routing:} The routing function selects a video generator based on embodiment dynamics (e.g., single-image diffusion for humanoid viewpoints, or keyframe-interpolation diffusion for drone dynamics).
\item \textbf{Prompt Construction:} A vision-language model produces an enhanced prompt describing only dynamic ego-motion changes relative to $I_0$, explicitly avoiding re-description of static scene elements to prevent hallucinated geometry.
\item \textbf{Video Generation:} The selected diffusion model synthesizes a short first-person navigation clip (5--15\,s).
\item \textbf{Structured Validation:} A reasoning-capable evaluator critiques the video, returning scalar scores for PA, PP, and VQ, alongside a structured critique identifying dominant failure modes (e.g., obstacle fixation, trajectory scale mismatch).
\item \textbf{Bottleneck Refinement:} If $\mathrm{mean}(\text{PA},\text{PP},\text{VQ}) < \tau$, the agent identifies the dominant bottleneck and modifies the prompt or pipeline accordingly.
\end{itemize}

\subsubsection{Three-Tier Memory Architecture}
To improve cross-task reliability and reduce iteration count, Stage~I maintains three forms of memory:
\paragraph{Static Constitutional Rules.}
A fixed rule set (\texttt{rules.md}) enforces behavioral constraints such as bottleneck-first optimization, anti-hallucination guardrails, and structured output formatting.
\paragraph{Long-Term Memory ($M_{LT}$).}
A persistent store that accumulates successful strategies across tasks, logging effective model--embodiment pairings, negative prompt templates, and recovery strategies for tight passages.
\paragraph{Short-Term Memory ($M_{ST}$).}
A per-task buffer that records iteration history and failed refinement attempts, preventing cyclic repetition of ineffective strategies.

\subsubsection{Output Representation}
The final output of Stage~I is the validated first-person navigation video $V_{goal}$. This video serves as a high-bandwidth \emph{Visual Intermediate Representation~(VIR)} that explicitly encodes:
\begin{itemize}
    \item Directional ego-motion and trajectory curvature.
    \item Local geometry for relative obstacle avoidance.
    \item Temporal stopping behavior dictated by the language instruction.
\end{itemize}
Importantly, by resolving semantic ambiguity in the pixel domain, the VIR decouples high-level reasoning from low-level control, providing a dense, embodiment-aware prior for Stage~II.

\subsection{Stage~II: Flow-Constrained Diffusion Transformer (FlowDiT)}

\begin{figure}[t]
    \centering
    \includegraphics[width=\columnwidth]{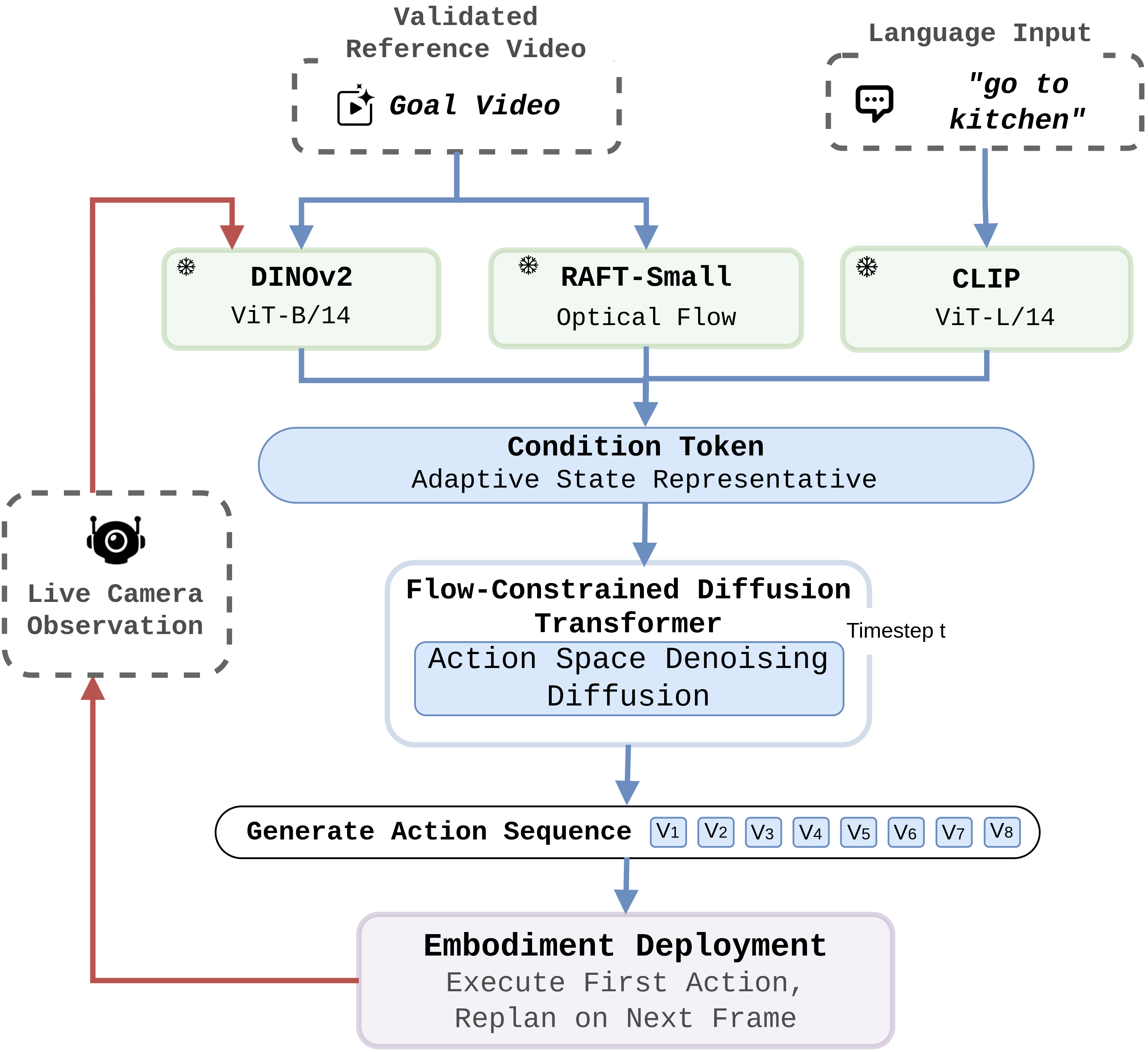}
    \caption{Stage~II: FlowDiT execution module. FlowDiT conditions on the validated reference video $V_{\mathrm{goal}}$ (visual features and motion cues) and the instruction $L$ to predict a horizon-$H$ sequence of velocity commands $(v_x, v_y, \omega)$. The policy executes in a receding-horizon manner by applying the first action and re-evaluating at the next timestep (with optional live observation $I_t$ when available).}
    \label{fig:stage2}
    \vspace{-0.5cm}
\end{figure}

Stage~II converts the validated reference video $V_{goal}$
into continuous velocity commands in a real-time closed-loop
control setting. Unlike classical trajectory tracking methods
that operate on explicit geometric paths, FlowDiT learns a
conditional action distribution directly from visual
and motion representations (see Fig.~\ref{fig:stage2}).

\subsubsection{Problem Formulation}

Given:
\begin{itemize}
\item A validated goal video $V_{goal}$ from Stage~I,
\item A natural language instruction $L$,
\item Optionally, a live observation $I_t$ (omitted in open-loop mode),
\end{itemize}

FlowDiT predicts a horizon-$H$ sequence of velocity commands:

\begin{align}
a_{t:t+H-1} =
\{(v_x, v_y, \omega)\}_{t:t+H-1},
\quad H=8,
\end{align}
where $v_x$ and $v_y$ denote translational velocities
in the robot frame and $\omega$ denotes angular velocity.

Only the first action in the predicted sequence
is executed. The model is re-evaluated at the next
timestep with updated observation $I_{t+1}$,
yielding a receding-horizon model-predictive
diffusion control scheme operating at 40--47\,Hz.

\subsubsection{Conditioning Representation}

FlowDiT is conditioned on a 2304-dimensional vector $c$
that integrates semantic, temporal, and linguistic cues:

\begin{align}
c =
[
\underbrace{\text{goal}_{\text{vision}}}_{768}
\parallel
\underbrace{\text{goal}_{\text{flow}}}_{256}
\parallel
\underbrace{\text{obs}_{\text{vision}}}_{768}
\parallel
\underbrace{\text{goal}_{\text{lang}}}_{512}
].
\end{align}

Each component is defined as follows:

\paragraph{Goal Vision Embedding.}
We encode selected frames of $V_{goal}$ using
DINOv2~\cite{dinov2_2023}, producing a 768-dimensional representation
capturing scene layout and semantic structure.

\paragraph{Goal Flow Embedding.}
A learned 6-channel CNN (5.2M parameters) extracts optical flow
between consecutive frames of $V_{goal}$.
The resulting flow fields are aggregated through
temporal pooling and projected to a 256-dimensional
embedding.
This term provides explicit ego-motion cues,
disambiguating approach versus retreat
and stabilizing short-horizon control.

\paragraph{Live Observation Embedding (Optional).}
When available, the current camera frame $I_t$ is encoded using
DINOv2 to produce a 768-dimensional state representation.
In open-loop mode, this term is replaced by a repeated encoding of the last frame of $V_{goal}$.

\paragraph{Language Embedding.}
The instruction $L$ is encoded using CLIP~\cite{clip_2021}
to produce a 512-dimensional semantic vector.
This term enables language-conditioned stopping
and disambiguation of visually similar goals.

\subsubsection{Diffusion Policy Formulation}

We model the conditional action distribution
using a denoising diffusion process.
Let $a_0 \in \mathbb{R}^{H \times 3}$ denote the
ground-truth action block.
We sample Gaussian noise $\epsilon \sim \mathcal{N}(0,I)$
and define the forward diffusion process:

\begin{align}
a_t =
\sqrt{\bar{\alpha}_t} a_0 +
\sqrt{1-\bar{\alpha}_t}\,\epsilon,
\end{align}
where $\bar{\alpha}_t$ is the cumulative product
of $(1-\beta_t)$ under a linear noise schedule
$\beta_1=10^{-4}$ to $\beta_T=2\times10^{-2}$
with $T=100$.

The model $\epsilon_\theta$ predicts the injected noise:

\begin{align}
\mathcal{L} =
\mathbb{E}_{a_0,t,\epsilon}
\left[
\left\|
\epsilon -
\epsilon_\theta(a_t, t, c)
\right\|^2
\right].
\end{align}
Inference uses deterministic DDIM sampling~\cite{ddim_2021}
with 10 denoising steps,
balancing stability and latency.

\subsubsection{Transformer Architecture}

FlowDiT adopts a Diffusion Transformer~(DiT)
architecture~\cite{dit_2023} with
8 transformer blocks,
hidden dimension 512,
8 attention heads,
and horizon tokenization over $H=8$ timesteps.

Each action timestep is embedded as a token.
Conditioning is injected through
\textbf{adaLN-Zero} modulation.

In each block, the timestep embedding $t$
and condition vector $c$
are processed through an MLP:

\begin{align}
(\gamma, \beta, g) = \mathrm{MLP}(c, t),
\end{align}
which modulates LayerNorm activations as:

\begin{align}
\mathrm{LN}(h) \rightarrow
g \cdot (\gamma \odot \mathrm{LN}(h) + \beta).
\end{align}

This conditioning mechanism avoids concatenating
$c$ to every token and preserves architectural efficiency.

\subsubsection{Closed-Loop Execution and Stability}

During deployment, FlowDiT supports both \emph{closed-loop} and \emph{open-loop} execution. In both modes, it operates in a model-predictive manner: (1)~encode $V_{goal}$ and (if available) $I_t$; (2)~predict action block $a_{t:t+7}$; (3)~execute only $a_t$; (4)~advance to $t+1$ and repeat. In open-loop mode, $I_t$ is omitted and the policy relies entirely on the reference video and language conditioning, enabling deployment on platforms where streaming camera feedback is impractical.

This receding-horizon strategy stabilizes control
under visual perturbations and minor trajectory drift.
Empirically, the system runs at $\sim$20\,ms per step
on an RTX~5090,
achieving 40--47\,Hz closed-loop frequency.

\subsubsection{Parameter Efficiency}

FlowDiT leverages a frozen DINOv2 encoder
(86.6M parameters) and a frozen CLIP encoder,
training only 43M parameters within the DiT module
(including the 5.2M learned flow encoder).
This separation maintains representation richness
while keeping trainable capacity compact,
yielding a $161\times$ reduction relative to
end-to-end VLA models with billions of parameters.

\section{Experimental Setup}

\subsection{Task Suite and Embodiments}

We evaluate Action Agent across 50 first-person navigation tasks
in indoor environments including warehouses and hospital corridors. Each task consists of a natural language instruction, an initial first-person observation, and a predefined success condition based on goal proximity and stable stopping.

The task suite includes 15 straight-line tasks,
16 obstacle-avoidance tasks, 10 path-following tasks,
and 9 orientation or turning tasks.

\begin{figure}[t]
    \centering
    \includegraphics[width=\columnwidth]{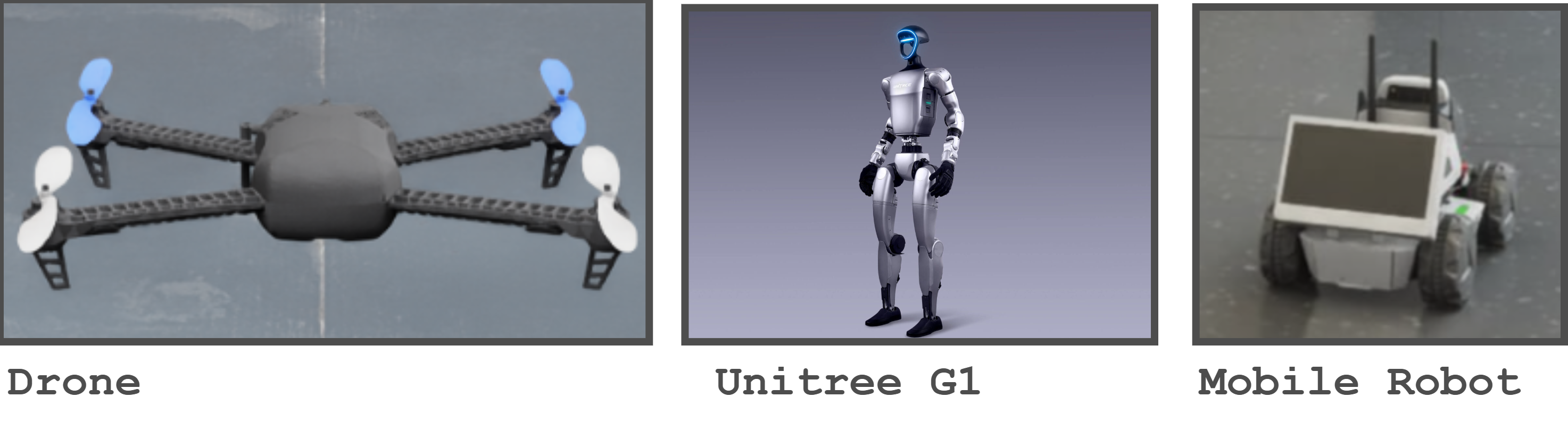}
    \caption{Robot embodiments used for FlowDit evaluation: a quadrotor drone (Isaac Sim), a Unitree G1 humanoid (real hardware), a wheeled mobile robot (Isaac Sim).
    }
    \label{fig:robot}
    \vspace{-0.5cm}
\end{figure}

To evaluate cross-embodiment generalization,
we deploy Action Agent on three distinct robotic platforms.The three evaluation embodiments are shown in Fig.\ref{fig:robot}:

\begin{itemize}
\item \textbf{Unitree G1 Humanoid (Real Hardware):} We deploy Action Agent on a Unitree G1 humanoid using a head-mounted RGB camera (approximately 1.2\,m height) to capture the initial observation $I_0$ for Stage~I. The resulting velocity commands are executed on the robot \emph{open-loop} (i.e., without live RGB feedback during motion). We evaluate task completion across 17 trials in unseen indoor lab/office environments. Success is defined as reaching the intended goal region or completing the traversal without obstacle collision; 11 of 17 trials succeed (\textbf{64.7\%}).
\item \textbf{Quadrotor Drone (Isaac Sim):}
13 tasks involving forward motion,
altitude adjustments, and banking turns.

\item \textbf{Wheeled Mobile Robot (Isaac Sim):}
12 tasks with planar differential drive kinematics.
\end{itemize}

All embodiments use the same Stage~I video generation
pipeline and the same FlowDiT controller,
without retraining per embodiment.
\subsection{Evaluation Metrics}

We evaluate both Stage~I (video synthesis) and Stage~II (navigation execution).

\paragraph{Stage~I Metrics.}
Each generated reference video is scored by the evaluator:

\begin{itemize}
\item \textbf{Prompt Adherence (PA)}: alignment with instruction.
\item \textbf{Physical Plausibility (PP)}: motion realism and collision avoidance.
\item \textbf{Visual Quality (VQ)}: temporal and perceptual coherence.
\end{itemize}

A video is considered valid if
$\mathrm{mean}(\text{PA},\text{PP},\text{VQ}) \ge 80$.

\paragraph{Stage~II Metrics.}
For simulation, we report success rate~(SR), mean Absolute Trajectory Error~(ATE), and Direction Accuracy~(DA). For real-robot experiments, we report task completion rate across trials in unseen environments.

\subsection{Training}

FlowDiT follows a two-stage training strategy. We first pretrain on the RECON dataset~\cite{recon_2021} from the Open X-Embodiment collection, which provides 11,830 outdoor navigation episodes collected on a Clearpath Jackal wheeled robot with scripted exploration. This pretraining stage allows FlowDiT to learn general visual navigation priors, like scene understanding, obstacle avoidance patterns, and ego-motion representation, from a large and diverse outdoor dataset.

We then fine-tune on Unitree G1 humanoid navigation episodes (162 train / 41 val) collected in Isaac Sim indoor environments (warehouse and hospital corridors). This fine-tuning stage calibrates the velocity output dynamics to match the G1 humanoid's motion characteristics, enabling accurate video-to-action velocity transfer for real-world deployment. The fine-tuning configuration is summarized in Table~\ref{tab:training_summary}.

\begin{table}[t]
\centering
\vspace{0.4cm}
\caption{Fine-Tuning Configuration (G1 Sim $\rightarrow$ Real)}
\label{tab:training_summary}
\begin{tabular}{ll}
\toprule
Episodes & 203 (162 train / 41 val) \\
Frame Resolution & $224\times224$ RGB \\
Action Horizon & $H=8$ \\
Action Space & $[-1,1]^3$ normalized velocities \\
Optimizer & AdamW (LR $1\mathrm{e}{-4}$) \\
Batch Size & 8 (FP16) \\
Noise Schedule & Linear $\beta_1\!=\!10^{-4}$ to $\beta_T\!=\!2\!\times\!10^{-2}$ \\
Diffusion Steps & 100 (train) / 10 (DDIM infer) \\
Trainable Params & $\approx$43M \\
Hardware & NVIDIA RTX~5090 (32\,GB) \\
\bottomrule
\end{tabular}
\end{table}

\subsection{Baselines}

We compare FlowDiT against representative
navigation baselines. We note that published baseline numbers are evaluated on the RECON outdoor dataset with different robots and success criteria; the comparison is for \emph{architectural context} rather than direct benchmarking.

\begin{itemize}
\item \textbf{ViNT}~\cite{vint_2023} (vision transformer navigation)
\item \textbf{NoMaD}~\cite{nomad_2024} (goal-conditioned diffusion navigation)
\item \textbf{OpenVLA}~\cite{openvla_2024} (foundation VLA model)
\item \textbf{Vision-only Diffusion} (ours, no optical flow)
\end{itemize}

\section{Results}

\subsection{Stage~I: Agentic Synthesis Reliability}
The introduction of an agentic optimization loop significantly improves the reliability of navigation video synthesis. While single-shot generation without optimization succeeds in only 35\% of tasks, the full Action Agent framework achieves an 86\% overall success rate across all embodiments, as shown in Table~\ref{tab:stage1_results}.

\begin{table}[h]
\centering
\caption{Trajectory Synthesis Performance by Embodiment}
\label{tab:stage1_results}
\begin{tabular}{lcccc}
\toprule
Embodiment & Pass & PA & PP & VQ \\
\midrule
Unitree G1 (Real) & 92\% & 92.3 & 91.5 & 93.1 \\
Drone (Sim) & 77\% & 84.6 & 82.3 & 88.1 \\
Mobile Robot (Sim) & 83\% & 87.5 & 85.8 & 89.2 \\
\bottomrule
\end{tabular}
\end{table}

\subsubsection{Convergence and Memory Impact}
The agentic optimization loop demonstrates steady convergence, with mean scores (PA, PP, VQ) progressing from 65.0 at iteration~1 to 87.0 at iteration~5. The inclusion of long-term memory is critical; it reduces the average number of iterations from 4.8 to 3.5 while increasing the overall success rate by 14\%.

\subsection{Stage~II: Navigation Performance}

\subsubsection{Baseline Comparisons}
We evaluate FlowDiT against state-of-the-art navigation and Vision-Language-Action~(VLA) baselines. As detailed in Table~\ref{tab:baselines}, FlowDiT achieves a 73.2\% success rate in simulation with $161\times$ fewer trainable parameters than OpenVLA~\cite{openvla_2024}. Although datasets and environments differ across methods (see Table ~\ref{tab:baselines} caption), the comparison demonstrates that conditioning a lightweight diffusion policy on explicit visual and motion representations is competitive with end-to-end, billion-parameter models.

\begin{table}[h]
\centering
\vspace{0.4cm}
\caption{Architectural Comparison. $\dagger$Published on RECON (outdoor, different robots/criteria). $\ddagger$Measured on G1 val (41 episodes, simulation, with post-processing).}
\label{tab:baselines}
\begin{tabular}{lcccl}
\toprule
Method & Flow & Params & SR\,(\%) & Data \\
\midrule
ViNT~\cite{vint_2023} & No & $\sim$25M & 61$\dagger$ & RECON \\
NoMaD~\cite{nomad_2024} & No & $\sim$100M & 68$\dagger$ & RECON \\
OpenVLA~\cite{openvla_2024} & No & $\sim$7B & 65$\dagger$ & Mixed \\
\textbf{FlowDiT (Ours)} & \textbf{Yes} & \textbf{43M} & \textbf{73.2$\ddagger$} & \textbf{G1 sim} \\
\textit{FlowDiT (real)} & \textit{Yes} & \textit{43M} & \textit{64.7} & \textit{G1 real} \\
\bottomrule
\end{tabular}
\end{table}

\subsubsection{Real-Hardware Transfer (Unitree G1)}
To evaluate zero-shot transferability, we deployed Action Agent on a physical Unitree G1 in unseen indoor lab/office environments. Each trial followed the full pipeline: capture $I_0$ from the robot's camera, generate a navigation video via Stage~I given a language instruction (e.g., ``walk toward the table'', ``Maps to the corridor end'', ``move past the shelf and turn right''), extract velocity commands via FlowDiT, and execute open-loop (no RGB feedback). Success is task completion: reaching the goal region or collision-free traversal. Across 17 trials, \textbf{11 succeeded (64.7\%)}.

The six failures stem from three open-loop-specific modes: (i)~\textbf{video-to-metric scale ambiguity} (most frequent) the generated video implies a motion magnitude mismatched with actual robot displacement; (ii)~\textbf{trajectory divergence} small heading errors compound without correction; (iii)~\textbf{obstacle collision} from accumulated drift. These do not reflect errors in the generated trajectory: Stage~I consistently produces topologically correct paths. Closed-loop execution, conditioning on live $I_t$ at 40--47\,Hz for online correction and visual arrival detection, is expected to address all three modes.

\subsection{Ablation Study: Disentangling Modality Contributions}
To understand the specific contributions of our multi-modal conditioning, we conducted an ablation study on the G1 simulation validation split (41 episodes, with post-processing). All variants use the same trained checkpoint; individual modality features are zeroed at inference to isolate their contribution. Results are shown in Table~\ref{tab:ablation}.

\begin{table}[t]
\centering
\caption{Ablation study on modality contributions. G1 val split (41 episodes), with post-processing. All variants use the same checkpoint; features zeroed at inference.}
\label{tab:ablation}
\begin{tabular}{llccc}
\toprule
Variant & Components & SR\,(\%) & ATE\,(m) & DA\,(\%) \\
\midrule
Vision-only & DINOv2 & 58.5 & 0.298 & 73.0 \\
No flow & DINOv2{+}CLIP & 73.2 & 0.309 & 67.8 \\
No language & DINOv2{+}Flow & 65.9 & 0.280 & 80.6 \\
\textbf{Full FlowDiT} & \textbf{All three} & \textbf{73.4} & \textbf{0.293} & \textbf{76.1} \\
\midrule
\multicolumn{2}{l}{\textit{Real robot (17 trials, open-loop)}} & \textit{64.7} & --- & --- \\
\bottomrule
\end{tabular}
\end{table}

\textbf{Flow improves direction accuracy:} Removing the learned flow encoder drops Direction Accuracy from 76.1\% to 67.8\% ($-$8.3\%), confirming that optical flow is critical for directional disambiguation. Without flow, FlowDiT relies solely on semantic vision, which struggles with ``looming'' depth ambiguities and estimating the temporal proximity of approaching obstacles.

\textbf{Language enables goal discrimination:} Removing the CLIP language embedding reduces SR from 73.4\% to 65.9\%. In these trials, FlowDiT successfully followed the visual trajectory of the reference video but frequently failed to execute \emph{semantic stopping}, either overshooting the goal or stopping at incorrect landmarks.

\textbf{Vision-only baseline:} Using only DINOv2 features (no flow, no language) yields the lowest SR at 58.5\%, demonstrating that both modalities provide complementary information essential for robust navigation.

\subsection{System Efficiency and High-Frequency Execution}
A critical limitation of end-to-end VLA models is their immense parameter count (e.g., 7B for OpenVLA), which precludes high-frequency control on standard hardware. FlowDiT generates 121 velocity waypoints from a single reference video in 3.68\,s average inference time on an RTX~5090, running at 40--47\,Hz ($\sim$20\,ms/step). The 43M trainable parameters represent a $161\times$ reduction vs.\ OpenVLA. Post-processing (EMA smoothing, velocity clamping, yaw scaling) is applied to diffusion outputs before execution.

\section{Discussion}

Our results demonstrate explicitly decoupling trajectory imagination from low-level execution yields substantial improvements in cross-embodiment transfer. Generative video, refined through an agentic loop, acts as a universal, high-bandwidth \emph{Visual Intermediate Representation~(VIR)} between semantic reasoning and metric control. Furthermore, explicitly injecting ego-motion constraints via learned optical flow resolves the ``looming'' depth ambiguities that static semantic encoders struggle to process, while the diffusion policy naturally captures multi-modal action distributions.

\textbf{Video-to-Action Transfer:} The two-stage training strategy (pretraining on RECON's large-scale outdoor data followed by fine-tuning on 203 G1 simulation episodes) enables effective video-to-action transfer. The RECON pretraining provides general navigation priors, while the Isaac Sim fine-tuning calibrates velocity dynamics to the humanoid embodiment. This calibration is critical: the real-robot success rate of 64.7\% on task completion in unseen environments demonstrates that the sim-to-real velocity transfer is viable, with a gap of only $\sim$8.5\% relative to simulation.

\textbf{Hardware Transfer and the Metric-Semantic Gap:} The open-loop hardware study revealed the \emph{video-to-metric scale ambiguity}: without online observation feedback, minor mismatches between generated pixel-space motion and real-world displacement accumulate into trajectory drift. Notably, FlowDiT still succeeds in the majority of open-loop trials, demonstrating that the reference video alone carries substantial navigational information. Engaging the closed-loop variant, conditioning on live observations $I_t$, is expected to further improve robustness by dynamically grounding semantic imagination into metric reality.

\textbf{Controllers for World Models:} As foundation video models scale into general-purpose ``World Models'' capable of simulating complex physics, they still lack native interfaces for physical robotic embodiment. Action Agent bridges this gap. By utilizing an LLM as a meta-optimizer in Stage~I, we provide a mechanism to steer and validate these world models for specific task constraints. As generative video models improve in latency and physical realism, the Action Agent architecture will directly inherit these capabilities, further reducing the sim-to-real gap without requiring retraining of the Stage~II control policy.

Despite these advantages, the system exhibits some limitations. FlowDiT experiences elevated failure rates in highly constrained spaces, such as narrow doorframes, where centimeter-level precision is required. Additionally, current video generation architectures limit clip durations to 5--15\,s, constraining the horizon of single-stage planning.

\section{Conclusion}

We introduced Action Agent, a two-stage framework that unifies agentic video synthesis with flow-constrained diffusion for embodied robot navigation. Pretrained on RECON and fine-tuned on 203 Unitree G1 simulation episodes to calibrate velocity dynamics, a single 43M-parameter FlowDiT checkpoint achieves 73.2\% navigation success in simulation and 64.7\% task completion on a real Unitree G1 humanoid in unseen indoor environments under open-loop execution. Action Agent provides a highly efficient, embodiment-aware paradigm for grounding visual foundation models into physical action.
 
 Future work explores receding-horizon video generation, closed-loop hardware validation, and aerial training data.

\balance

\end{document}